\def\year{2022}\relax
\documentclass[letterpaper]{article} 
\usepackage{aaai22}  
\usepackage{times}  
\usepackage{helvet} 
\usepackage{courier}  
\usepackage[hyphens]{url}  
\usepackage{graphicx} 
\usepackage{natbib}  
\usepackage{caption} 
\usepackage[ruled, linesnumbered]{algorithm2e}
\usepackage{amsfonts}
\usepackage{amsmath}
\usepackage{booktabs}
\usepackage{diagbox}
\usepackage{graphicx}
\usepackage{multirow}
\usepackage{subfigure}
\usepackage{xcolor}
\title{Attacking Video Recognition Models with Bullet-Screen Comments}
\author{
    Kai Chen\textsuperscript{\rm 1,\rm 2},
    Zhipeng Wei\textsuperscript{\rm 1,\rm 2},
    Jingjing Chen\equalcontrib\textsuperscript{\rm 1,\rm 2}
    Zuxuan Wu\textsuperscript{\rm 1,\rm 2},
    Yu-Gang Jiang\equalcontrib\textsuperscript{\rm 1,\rm 2}\\
}
\affiliations{
    \textsuperscript{\rm 1}Shanghai Key Lab of Intelligent Information Processing, School of Computer Science, Fudan University\\
    \textsuperscript{\rm 2}Shanghai Collaborative Innovation Center on Intelligent Visual Computing\\
    \{kaichen20, chenjingjing, zxwu, ygj\}@fudan.edu.cn, zpwei21@m.fudan.edu.cn\\
}

\begin{document}

\maketitle
\begin{abstract}
Recent research has demonstrated that Deep Neural Networks (DNNs) are vulnerable to adversarial patches which introduce perceptible but localized changes to the input. Nevertheless, existing approaches have focused on generating adversarial patches on images, their counterparts in videos have been less explored. Compared with images, attacking videos is much more challenging as it needs to consider not only spatial cues but also temporal cues. To close this gap, we introduce a novel adversarial attack in this paper, the bullet-screen comment (BSC) attack, which attacks video recognition models with BSCs. Specifically, adversarial BSCs are generated with a Reinforcement Learning (RL) framework, where the environment is set as the target model and the agent plays the role of selecting the position and transparency of each BSC. By continuously querying the target models and receiving feedback, the agent gradually adjusts its selection strategies in order to achieve a high fooling rate with non-overlapping BSCs. As BSCs can be regarded as a kind of meaningful patch, adding it to a clean video will not affect people’s understanding of the video content, nor will arouse people’s suspicion. We conduct extensive experiments to verify the effectiveness of the proposed method. On both UCF-101 and HMDB-51 datasets, our BSC attack method can achieve about 90\% fooling rate when attacking three mainstream video recognition models, while only occluding \textless 8\% areas in the video. Our code is available at \url{https://github.com/kay-ck/BSC-attack}.
\end{abstract}

\section{Introduction}
Deep Neural Networks (DNNs) have demonstrated superior performance in various video-related tasks \cite{song2021spatial,su2020video,han2021fine,wang2021visual}, like video recognition \cite{karpathy2014large,carreira2017quo,wu2016multi,zhang2021videolt}, video caption \cite{yang2017catching,liu2020sibnet} and video segmentation \cite{nilsson2018semantic,wang2019learning}, etc. However, recent works have shown that DNNs are extremely vulnerable to video adversarial examples which are generated by applying negligible perturbations to clean input samples \cite{wei2019sparse}. The existence of video adversarial examples leads to security concerns of Deep Learning-based video models in real-world applications. Therefore, it has attracted increasing research interest in recent years \cite{wei2021boosting,wei2021crossmodal,wei2020heuristic}.

\begin{figure}[t]
  \centering
  \includegraphics[width=0.75\linewidth]{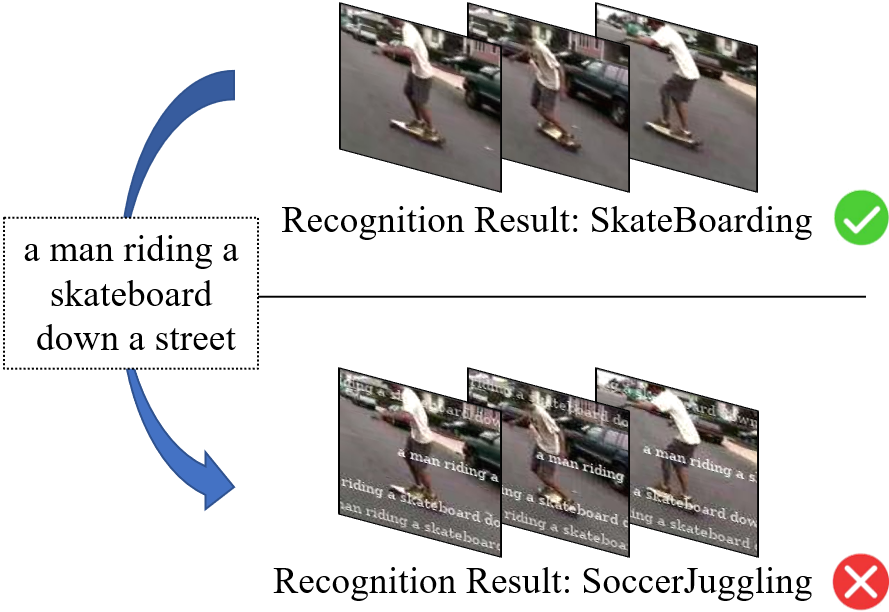}
  \caption{An illustration of adversarial BSC attacks. Given a video, it can successfully fool the video recognition model by adding BSCs.}
  \label{fig-vis}
\end{figure}

Nevertheless, most of the existing works focus on perturbation-based attacks, which introduce imperceptible changes to the clean input samples. The perturbations are constrained to have a small $L_p$ norm and applied to the whole input. While perturbation-based attacks have been demonstrated to be effective in attacking the video recognition models, they are typically difficult to apply in the physical world. In contrast, patch-based attacks generate adversarial patches by modifying the pixels within a restricted region without any limitations on the range of changes. Therefore, patch-based attacks are stronger and more effective in the physical world. Nevertheless, existing works on patch-based attacks are mostly focused on images, patch-based attacks on videos have seldom been explored.

This paper investigates patch-based attacks on videos in the black-box setting, where the adversary can only access the output of the target model. The challenges of this task mainly come from two aspects. First, a video is a sequence of images on which the adjacent frames are closely correlated. If selecting several frames in the video as in the case of perturbation-based video attacks \cite{wei2020heuristic} to add adversarial patches, it will increase the perceptibility of the attack. Second, compared to images, the dimension of videos is much higher. If attaching adversarial patches to each frame of the video, it will significantly increase the computation cost. Hence how to efficiently generate inconspicuous adversarial patches for video models in the black-box setting is the main challenge.

To address the aforementioned challenges, we propose a novel adversarial bullet-screen comment (BSC) attack method against video recognition models. As BSCs are quite popular when viewers watch videos online, people will be less sensitive to such meaningful patches than the rectangular patches \cite{yang2020patchattack} used in patch-based image attacks. To make the BSCs attached to each video different from each other, we introduce an image captioning model to automatically generate BSCs. Then the position and transparency of adversarial BSCs are selected based on two objectives. First, it should achieve a high fooling rate by placing the BSCs with selected transparencies on the selected positions. Second, BSCs should not overlap with each other in order to avoid obscuring the details of the video significantly. To this end, motivated by PatchAttack \cite{yang2020patchattack}, we formulate the search over the position and transparency of BSCs as a Reinforcement Learning (RL) problem to find the optimal positions and transparencies efficiently, resulting in a query efficient attack. Specifically, in RL, we define the environment as the target model and the agent as the role of position and transparency selection. By continuously querying the target model and receiving the feedback, the agent gradually adjusts its selection strategies in order to achieve a high fooling rate and zero Intersection over Union (IoU) between different BSCs. Figure~\ref{fig-vis} shows an example of our adversarial BSC attack. As can be seen, the few BSCs do not affect our understanding of the video but fool the video recognition model successfully.

\begin{figure*}[htbp]
  \centering
  \includegraphics[width=0.75\linewidth]{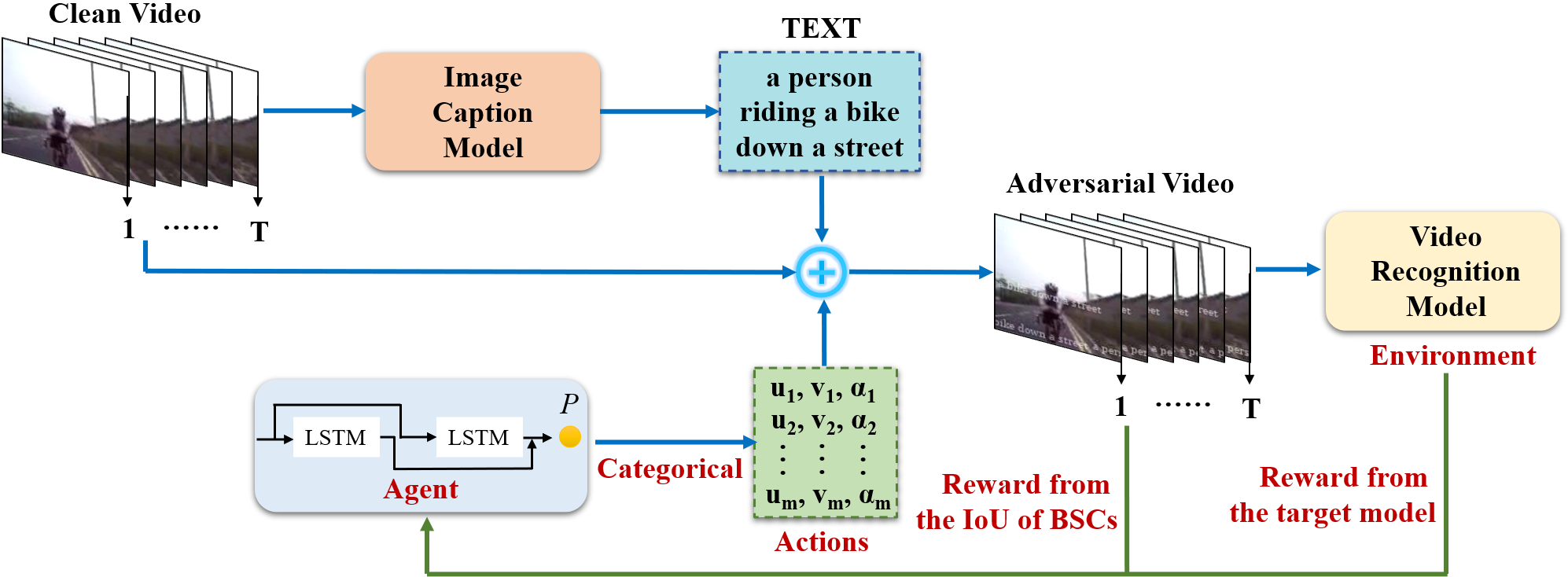}
  \caption{Overview of our black-box adversarial BSC attack method. We formulate the position, transparency selection and attacking step into an end-to-end RL framework.}
  \label{fig-overview}
\end{figure*}

Figure~\ref{fig-overview} overviews our proposed attack framework. Given a clean video sample, the content of BSCs is generated by an image captioning model. Then the position and transparency of BSCs are optimized through RL, where the agent adjusts the positions and transparencies according to two rewards (fooling rate and IoU between different BSCs) received from the environment (target model). By continuously querying the target model, the optimal positions and transparencies are selected to generate the video adversarial example. For the agent, we use a combination of a Long-Short Term Memory network (LSTM) and a fully connected (FC) layer.
In summary, our major contributions are as follows:
\begin{itemize}
\item We propose a novel BSC attack method against video recognition models. By formulating the attacking process with RL, our attack method achieves an efficient query.
\item We design a novel reward function that considers the IoU between BSCs to ensure that the added few BSCs do not affect the understanding of videos.
\item Extensive experiments on three widely used video recognition models and two benchmark video datasets (UCF-101 and HMDB-51) show that our proposed adversarial BSC framework can achieve high fooling rates.
\end{itemize}

\section{Related Work}
In this section, we provide a short review of perturbation-based attacks on video models and patch-based attacks.

\subsection{Perturbation-based Attacks on Video Models}
Perturbation-based attacks introduce imperceptible changes to the input that are restricted to have a small $L_p$ norm and are typically applied to the whole. Perturbation-based attacks on image models are firstly explored by Szegedy et al. \cite{szegedy2013intriguing}, where they add some imperceptible noises on clean images and mislead well-trained image classification models successfully. Sparked by this work, perturbation-based attacks on image models have been extensively studied \cite{goodfellow2014explaining,madry2017towards,carlini2017towards,chen2017zoo,ilyas2018black,wei2021towards,shi2021decision}. In the past years, perturbation-based attacks have been extended to video models. In terms of white-box attacks, where the adversary has complete access to the target model such as model parameters, model structure, etc, \cite{wei2019sparse} first proposes an $L_{2,1}$ norm regularization-based optimization algorithm to compute sparse adversarial perturbations for videos. \cite{li2019stealthy} leverages Generative Adversarial Network (GAN) to generate universal perturbations offline against real-time video classification systems, and the perturbations work on unseen inputs. \cite{chen2021appending} proposes to append a few dummy frames to a video clip and then add adversarial perturbations only on these new frames. For black-box attacks, \cite{jiang2019black} first utilizes tentative perturbations transferred from the image classification model and partition-based rectifications estimated by the Natural Evolutionary Strategies to obtain good adversarial gradient estimates with fewer queries to the target model. To boost the attack efficiency and reduce the query numbers, \cite{wei2020heuristic} proposes to heuristically search a subset of frames and adversarial perturbations are only generated on the salient regions of selected frames. More recently, \cite{zhang2020motion} proposes a motion-excited sampler to generate sparked prior and obtain significantly better attack performance. However, black-box perturbation-based attacks often require lots of queries and are difficult to apply in the physical world.

\subsection{Patch-based Attacks}
Patch-based attacks superimpose adversarial patches onto a small region of the input to create the adversarial example, making the attack more effective and applicable in the physical world by breaking the $L_p$ norm limitations in perturbation-based attacks. At present, patch-based attacks are mainly focused on image models. Adversarial patches are first proposed by \cite{brown2017adversarial}, which fools image classification models to ignore other scenery semantics and make wrong predictions by superimposing a relatively small patch onto the image. \cite{fawzi2016measuring} introduces the first black-box attack, which searches the position and shape of rectangular patches using Metropolis-Hastings sampling. \cite{ranjan2019attacking} further extends adversarial patches to optical flow networks and shows that such attacks can compromise their performance. Although these existing adversarial patches have powerful attack ability, they are highly conspicuous. To make adversarial patches be more inconspicuous, \cite{liu2019perceptual} introduces GAN to generate visually more natural patches. \cite{jia2020adv} further proposes to camouflage malicious information as watermarks to achieve adversarial stealthiness. This approach assumes that people’s understanding of the image content is not affected by such meaningful perturbations and hence will not arouse people’s suspicion, which is the similar assumption our approach is based on. In contrast, we disguise the adversarial patches as BSCs to attack video recognition models. As BSCs are meaningful and quite common, people will be less sensitive to such type of adversarial patch.

\section{Methodology}
\subsection{Problem formulation}
We denote the video recognition model as a function $F(\cdot)$, where $\theta_F$ denotes the model parameters. Given a clean video sample $x\in X\subset\mathbb{R}^{T\times W\times H\times C}$, where $X$ is the video space, $T$, $W$, $H$, $C$ denote the number of frames, frame width, frame height, and the number of channels respectively. For $x$, the associated ground-truth label $y\in Y=\{1,2,...,K\}$, where $Y$ is the label space, $K$ denotes the number of classes. We use $F(x):X \rightarrow Y$ to denote the prediction of the video recognition model $F(\cdot)$ for an input video $x$. The goal of adversarial attacks on video models is to generate an adversarial video $x_{adv}$ that can fool the video recognition model. There are two types of adversarial attacks: untargeted attacks and targeted attacks. Untargeted attacks make $F(x_{adv}) \neq y$, while targeted attacks make $F(x_{adv}) = y_{adv}$, where $y_{adv} \neq y$. In the case of untargeted attacks, we optimize the following objective function:
\begin{equation}
  \mathop{\arg}\mathop{\min}_{x_{adv}}\ \ -l(\bold{1}_{y},F(x_{adv})).
\end{equation}
where $\bold{1}_{y}$ is the one-hot encoding of the ground truth label, $l(\cdot)$ is the loss between the prediction and the ground truth label. In perturbation-based attacks, $x_{adv}$ is generated by modifying each pixel of the clean video, and the modification is constrained to have a small $L_p$ norm. In contrast, the only constraint for patch-based attacks is that the modification must be confined to a small region. 

In our work, we disguise adversarial patches as meaningful BSCs to achieve stealthiness. Specifically, the BSCs are confined to a sequence of regions within the video frames $\epsilon=\{\epsilon_{1},...,\epsilon_{t},...,\epsilon_{T}\}$, where $\epsilon_{t}$ denotes the region of BSCs (i.e., the set of pixels belonging to the region of BSCs) in the $t$-th frame. $\epsilon_{t}$ can be determined by giving the horizontal coordinate $u$ and vertical coordinate $v$ of the BSC's position in the first frame, the font size $h$, and the font type $\mathbb{T}$. Hence, the process of determining the $i$-th BSC's region in the first frame can be formalized as $\epsilon^{i}_{1} = R(TEXT,u_i,v_i,h,\mathbb{T}), i\in\{1,...,m\}$, where $R(\cdot)$ is the function that determines the region of BSCs in the video frames, $TEXT$ is the content of BSCs generated by the image captioning model, $m$ is the number of BSCs. To implement the BSCs floating from right to left across the video, we translate $\epsilon_{t}$ along the horizontal axis to get the region of BSCs in the $t+1$-th frame. Thus, we have $\epsilon^{i}_{t+1}=R(TEXT,u_i-t,v_i,h,\mathbb{T}), i\in\{1,...,m\}$.

To further mitigate the effect of BSCs on the video content, we use alpha blending in \cite{shen1998dct} to generate BSCs. When $(i,j)\in\epsilon_t$, the generation for $x_{adv}$ is formulated as:
\begin{equation}
  x_{adv}(t,i,j) = (p \ast \alpha + x(t,i,j) \ast (255 - \alpha)) / 255.
\end{equation}
On the contrary, when $(i,j) \notin \epsilon_t$, $x_{adv}$ is formulated as:
\begin{equation}
  x_{adv}(t,i,j) = x(t,i,j),
\end{equation}
where $(t,i,j)$ represents the position of the pixel in the video, $p$ represents the padding of the BSCs' region which is the color of the BSCs, $\alpha$ represents the value of the BSC's alpha channel which refers to the transparency of BSC's region w.r.t. the video background.

Note that in our paper, we only focus on optimizing the position and transparency of the BSC, instead of the color and rotation, etc.

\subsection{Position and Transparency Selection}
We use BSCs as adversarial patches, and the generation of video adversarial examples is only related to the position and transparency of BSCs. Searching over the position and transparency of BSCs can be formulated as an RL problem, since RL is demonstrated to be much more effective and efficient than random search strategies in \cite{yang2020patchattack}.

In the RL framework, the agent learns to select the position and transparency of adversarial BSCs by interacting with an environment that provides the rewards and updating its actions to maximize the total expected reward. In our work, the environment consists of $x$ and $F(\cdot)$, and an agent $\mathbb{A}$ is trained to sequentially search the position and transparency of BSCs. The searching space of BSCs' potential position and transparency is defined as:
\begin{gather}
  S=\{u_1,v_1,\alpha_1...,u_i, v_i,\alpha_i...,u_m,v_m,\alpha_m\},\notag
  \\u_i \in [-w, W], v_i \in [0, H-h], \alpha_i \in [127, 255].
\label{eqa:space}
\end{gather}
where $w$ is the width of the BSC, which depends on the content of the BSC. From Equation \ref{eqa:space}, it can be observed that $S$ has $3m$ dimensions, we set the agent $\mathbb{A}$ to take $3m$ actions in sequence to generate $a \in S$ and $a = \{a_1,...,a_{3m}\}$. Similar to \cite{yang2020patchattack}, we define the agent $\mathbb{A}$ to be a LSTM topped with a FC layer, its parameters are denoted by $\theta_{\mathbb{A}}$. The generation of actions is formulated as:
\begin{align}
  &a_0=0,\\
  &P=1,\\
  &\bold{h}_t=LSTM(\bold{h}_{t-1}, Embedding(a_{t-1})),\ t=\{1,...,3m\}.\\
  &p(a_t|(a_1,...,a_{t-1}))=softmax(\theta_W\times\bold{h}_t).\\
  &a_t=Categorical(p(a_t|(a_1,...,a_{t-1}))).\\
  &P=P \cdot p(a_t|(a_1,...,a_{t-1})).
\end{align}
where the initial input $a_0$ is set as 0, the hidden state $\bold{h}_t\in\mathbb{R}^{30}$ of LSTM evolves over step $t$, $\theta_W$ represents the weight of the FC layer. The FC layer that ends with the sigmoid function predicts the probability distribution $p(a_t|(a_1,...,a_{t-1}))$ over the possible actions for step $t$, and then one action $a_t$ are sampled via a $Categorical$ function and records the probability of the sampled action with $P$. The generated $a_t$ is fed back into LSTM in the next step, which drives the LSTM state transition from $\bold{h}_t$ to $\bold{h}_{t+1}$. This process is repeated until we have drawn a complete action of $3m$ steps.

To generate adversarial and non-overlapping BSCs, we define a reward that contains two components: the reward from the feedback of the target model $r_{attack}$ and the reward from the IoU between different BSCs $r_{IoU}$. The reward $r_{attack}$ and $r_{IoU}$ complement each other and work jointly to guide the learning of the agent:
\begin{equation}
  r=r_{attack}+\lambda \cdot r_{IoU}.
\end{equation}
The hyperparameter $\lambda$ is set according to the parameter tuning which will be discussed in Section \ref{sec:hp}. The former reward $r_{attack}$ makes the agent generate actions with a higher loss of the target model and is defined as:
\begin{equation}
  r_{attack}=log(1-\bold{1}_{y}\cdot F(x_{adv})).
\end{equation}
The reward $r_{IoU}$ avoids significantly obscuring the details of the video due to the overlap of BSCs, which is defined as:
\begin{equation}
  r_{IoU}=-IoU(\epsilon).
\end{equation}
$IoU(\cdot)$ calculates the intersection area over the union area between different BSCs. In this way, $r_{IoU}$ not only constrains the overlap between BSCs but also implicitly constrains the number of BSCs by regarding adversarial examples with overlapping BSCs as failures. Based on this reward, we expect the agent $\mathbb{A}$ to generate non-overlapping BSCs while successfully attack video recognition models.

Then, we employ the REINFORCE algorithm \cite{williams1992simple} to optimize the parameters $\theta_{\mathbb{A}}$ of the agent $\mathbb{A}$ by maximizing the expected reward $J(\theta_{\mathbb{A}})=E_P[r]$:
\begin{equation}
  \nabla_{\theta_{\mathbb{A}}}J(\theta_{\mathbb{A}})=\frac{1}{B}\sum_{n=1}^B\nabla_{\theta_{\mathbb{A}}}r_nlogP_n,
  \label{eq:loss}
\end{equation}
where $B$ is the batch size and is set as 500. We optimize the parameters via Adam with a learning rate of 0.03.
\begin{algorithm}
  \caption{Adversarial BSC attack}\label{algorithm1}
  \SetKwInOut{Input}{\textbf{Input}}
  \SetKwInOut{Parameter}{\textbf{Parameter}}
  \SetKwInOut{Output}{\textbf{Output}}
  \Input{video recognition model $F(\cdot)$, clean video $x$, ground-truth label $y$.}
  \Output{adversarial video $x_{adv}$.}
  \Parameter{the number of BSCs $m$, the font size $h$, the balancing factor $\lambda$, the font type $\mathbb{T}$.}
  \For{$i=1$ \KwTo $epochs$}
  {
  $TEXT = I(x[0])$ \;
  $a,P=\mathbb{A}(0)$ \;
  \For{$t=0$ \KwTo $T-1$}
  {
    $\epsilon^{m}_{t+1} = R(TEXT, u_i-t ,v_i, h, \mathbb{T}),i\in\{1,...,m\}$ \;
    \eIf{$(i,j)\in\epsilon_{t+1}$}{$x_{adv}(t+1,i,j)=(p\ast\alpha+x(t+1,i,j)\ast(255-\alpha))/255$}{$x_{adv}(t+1,i,j) = x(t+1,i,j)$}
  }
  $r_{attack}=log(1-\bold{1}_{y}\cdot F(x_{adv}))$ \;
  $r_{IoU}=-IoU(\epsilon)$ \;
  $r=r_{attack}+\lambda r_{IoU}$ \;
  Update the agent $\mathbb{A}$.
  }
  \Return $x_{adv}$
\end{algorithm}

\subsection{Overall Algorithm}
The overall process of our adversarial BSC attack is summarized in Algorithm \ref{algorithm1}. To enable automatically generate different BSCs for each video, a pre-trained image captioning model $I(\cdot)$ takes the first frame of clean video $x[0]$ as input and outputs the description that used as the BSC. Then, the agent generates an action sequence including position coordinates and transparency of $m$ BSCs, based on which the BSCs can be attached to the video and the rewards are calculated to optimize the agent finally. The attack process is repeated until we find the adversarial BSC with $r_{IoU}=0$, or the attack fails because the maximum query number is exceeded. Note that if there is more than one adversarial example with $r_{IoU}=0$ in the batch, we will select the one with the least salient region occluded by the BSCs. Intuitively, salient regions, for example, the foreground of the frames, have a high probability to be the human‘s focus area. Generating adversarial BSCs on the salient regions will be more likely to affect people’s understanding of the video content. Our approach is implemented on a workstation with four GPUs of NVIDIA GeForce RTX 2080 Ti. 

\section{Experiments}
\subsection{Experimental Setting}
\begin{table}[t]
    \centering
    \begin{tabular}{l c c c c} 
        \toprule
        $m$ & \multicolumn{1}{l}{$FR(\%)$} & \multicolumn{1}{l}{$AOA(\%)$} & \multicolumn{1}{l}{$AOA^\ast(\%)$} & \multicolumn{1}{l}{$AQN$} \\
        \midrule
        2 & 68.3 & 4.1 & 1.5 & 9084\\
        3 & 72.3 & 5.5 & 1.8 & 8089\\
        4 & 79.2 & 7.3 & 2.4 & 7005\\
        5 & 79.2 & 8.9 & 3.0 & 7292\\
        6 & 73.3 & 10.0 & 3.5 & 8233\\
        \bottomrule
    \end{tabular}
    \caption{Effects of the number of BSCs $m$.}
    \label{number_table}
\end{table}
\begin{table}[t]
    \centering
    \begin{tabular}{l c c c c}
        \toprule
        $h$ & \multicolumn{1}{l}{$FR(\%)$} & \multicolumn{1}{l}{$AOA(\%)$} & \multicolumn{1}{l}{$AOA^\ast(\%)$} & \multicolumn{1}{l}{$AQN$} \\ 
        \midrule
        7 & 78.2 & 7.1 & 2.3 & 7193\\
        8 & 79.2 & 7.3 & 2.4 & 7005\\
        9 & 80.2 & 7.6 & 2.5 & 6718\\
        10 & 81.2 & 8.4 & 2.8 & 6544\\
        11 & 82.1 & 9.2 & 3.0 & 6263\\
        12 & 81.2 & 10.6 & 3.7 & 6322\\
        13 & 76.2 & 10.6 & 3.8 & 7441\\
        \bottomrule
    \end{tabular}
    \caption{Effects of the font size $h$.}
    \label{size_table}
\end{table}
\begin{table}[t]
    \centering
    \begin{tabular}{l c c c c}
        \toprule
        $\lambda$ & \multicolumn{1}{l}{$FR(\%)$} & \multicolumn{1}{l}{$AOA(\%)$} & \multicolumn{1}{l}{$AOA^\ast(\%)$} & \multicolumn{1}{l}{$AQN$} \\
        \midrule
        $1e^{-5}$ & 79.2 & 7.7 & 2.6 & 7253\\
        $1e^{-4}$ & 80.2 & 7.8 & 2.5 & 6970\\
        $1e^{-3}$ & 80.2 & 7.6 & 2.5 & 6718\\
        $1e^{-2}$ & 78.2 & 7.5 & 2.6 & 7169\\
        $1e^{-1}$ & 76.2 & 7.8 & 2.6 & 7579\\
        \bottomrule
    \end{tabular}
    \caption{Effect of the balancing factor $\lambda$.}
    \label{factor_table}
\end{table}
\begin{table}[t]
    \centering
    \resizebox{.99\columnwidth}{!}{
    \begin{tabular}{l c c c c}
        \toprule
        $\mathbb{T}$ & \multicolumn{1}{l}{$FR(\%)$} & \multicolumn{1}{l}{$AOA(\%)$} & \multicolumn{1}{l}{$AOA^\ast(\%)$} & \multicolumn{1}{l}{$AQN$} \\
        \midrule
        $DejaVuSans$ & 78.2 & 7.5 & 2.4 & 7426\\
        $DejaVuSerif$ & 80.2 & 7.6 & 2.5 & 6718\\
        $DejaVuSansMono$ & 76.2 & 7.3 & 2.3 & 7753\\
        $DejaVuSansCondensed$ & 69.3 & 9.0 & 3.1 & 8797\\
        $DejaVuSerifCondensed$ & 67.3 & 8.8 & 3.0 & 9534\\
        \bottomrule
    \end{tabular}}
    \caption{Effect of the font type $\mathbb{T}$.}
    \label{type_table}
\end{table}
\textbf{Datasets.} We consider two popular benchmark datasets for video recognition: UCF-101 \cite{su2009ucf} and HMDB-51 \cite{kuehne2011hmdb}. UCF-101 is an action recognition dataset collected from YouTube, which contains 13,320 videos with 101 action categories. HMDB-51 is a dataset for human motion recognition and contains a total of 7000 clips distributed in 51 action classes. Both datasets split 70\% of the videos as training set and the remaining 30\% as test set. We randomly sample 2 videos from each category of the test dataset. During the test, 16-frame snippets are uniformly sampled from each video as input of target models. Note that, the sampled video snippet can all be classified correctly by target models.

\textbf{Target Models.} Three video recognition models, Long-term Recurrent Convolutional Network (LRCN) \cite{donahue2015long}, C3D \cite{hara2018can} and I3D-Slow \cite{feichtenhofer2019slowfast} are used as our target models. LRCN exploits the temporal information contained in successive frames, with Recurrent Neural Networks (RNNs) capturing long-term dependencies on the features generated by Convolutional Neural Networks (CNNs). In our implementation, Inception V3 \cite{szegedy2016rethinking} pre-trained on ImageNet is utilized to extract features from video frames and LSTM is utilized for video recognition; C3D applies 3D convolution to learn spatio-temporal features from videos with spatio-temporal filters for video recognition; I3D-Slow preserves the slow pathway, which operates at the low frame rate and captures spatial semantics in the SlowFast \cite{feichtenhofer2019slowfast} framework. These three models are the mainstream methods for video recognition. On UCF-101, the recognition accuracies for C3D, LRCN and I3D-Slow are 85.88\%, 64.92\% and 63.39\% respectively, while on HMDB-51, the recognition accuracies are 59.95\%, 37.42\% and 34.9\% respectively.

\textbf{Image Captioning Model.} For simplicity and efficiency, we adopt an attention-based image captioning model\cite{xu2015show} that is pre-trained on Microsoft Common Objects in Context (MS COCO) \cite{lin2014microsoft} to automatically generate the description for the first frame of videos.

\textbf{Metrics.} Three metrics are used to evaluate the performance of our method on various sides. 1) Fooling rate ($FR$): the ratio of adversarial videos that are successfully misclassified. 2) Average occluded area ($AOA$): the average area percentage occluded by BSCs in the entire video. We use $AOA^\ast$ to denote the average area percentage occluded by BSCs in the salient region. 3) Average query number ($AQN$): the average number of querying the target models to finish the attacks.

\subsection{Effects of Hyperparameters}\label{sec:hp}
We conduct a large number of experiments to determine four hyperparameters in Algorithm \ref{algorithm1}, including the number of BSCs $m$, the font size $h$, the balancing factor $\lambda$ in the reward, and the font type $\mathbb{T}$. We evaluate the attack performance of our algorithm on the C3D model with different hyperparameters. For the evaluation, we randomly sample 1 video per category from the test set of UCF-101. The sampled videos can be correctly classified by the C3D model. Then, we do a grid search to find the most appropriate values for these four hyperparameters.

Table~\ref{number_table} and Table~\ref{size_table} show the attack performance with different number of BSCs and different font sizes, respectively. The results show that when the number of BSCs $m$ increases, the $AOA$ will increase while the $FR$ will firstly increase and then decrease. When the font size $h$ increases, $AOA$ and $FR$ show a similar trend. That is, as the number of BSCs or the font size increases, more areas in the video are occluded, hence achieves a higher fooling rate. However, since we regard the adversarial examples with overlapping BSCs as failures, BSCs are more likely to overlap when the number of BSCs or the font size increases. To strike a balance between $FR$, $AOA$ and $AQN$, we set $m=4$ and $h=9$ to conduct subsequent experiments. Table~\ref{factor_table} shows the attack performance with different balancing factors in the reward. As can be seen from the table, when $\lambda$ increases, $FR$ decreases slightly while $AOA$ remains relatively stable. That is, when the reward $r_{IoU}$ has a larger weight, the model tends to make the generated BSCs non-overlap rather than optimize the attack success rate, hence results in a lower fooling rate. Therefore, we set $\lambda=1e^{-3}$ so that adversarial BSC attack can achieve the highest $FR$ (\%) and the least $AQN$. Table~\ref{type_table} shows the attack performance with different DejaVu font types. According to the results, we set $\mathbb{T}=DejaVuSerif$ to achieve the best attack performance for the adversarial BSC attack.

\subsection{Performance Comparison}
\begin{figure}[t]
  \centering
  \includegraphics[width=0.75\linewidth]{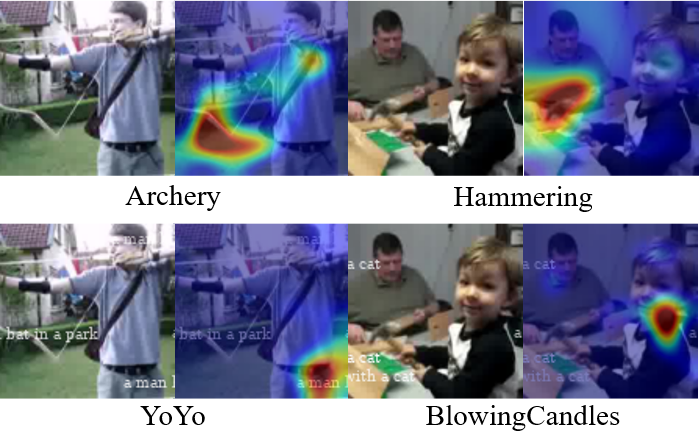}
  \caption{The top row is the clean frames and their corresponding heatmaps. The bottom row is the adversarial frames and their corresponding heatmaps.}
  \label{fig-heatmap}
\end{figure}
\begin{figure}[t]
  \centering
  \includegraphics[width=0.75\linewidth]{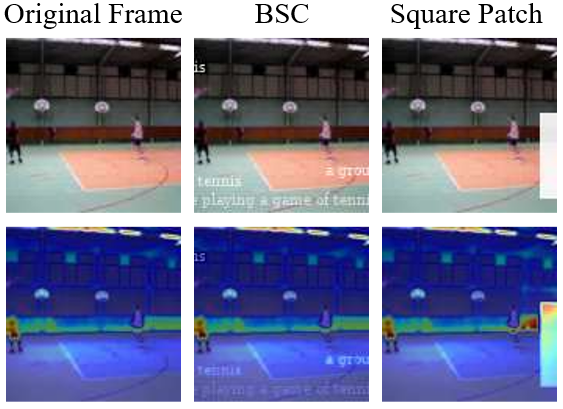}
  \caption{Examples of saliency detection for adversarial patches. We can see that our BSC does not trigger the saliency detection significantly.}
  \label{fig-saliencymap}
\end{figure}

We compare our method with PatchAttack\cite{yang2020patchattack}, which is originally proposed to attack image classification models in the black-box setting. Since BSCs are usually in white and untextured, for a fair comparison, we only consider the white square patch in the comparison. Different from the original setting of PatchAttack, we extend PatchAttack to attack video models by selecting the position and transparency of a white square patch with the same area as $m$ BSCs via RL. Besides, we also compare two variants of our method. One variant uses Basin hopping (BH) \cite{wales1997global} instead of RL to search over the position and transparency of BSCs. BH is a stochastic optimization algorithm that can be used to find the global minimum of a multivariate function. During each iteration, BH generates several new variables with random perturbation, then finds the local minimization, and finally accepts or rejects the new variables according to the minimized function value. The other variant randomly selects the position and transparency of the BSCs. For a fair comparison, we set the number of random trials equal to the query numbers of our method based on RL.

\begin{table*}[h]
    \centering
    \begin{tabular}{l c c c c c c}
        \toprule
        \multirow{2}{*}{Dataset} & \multirow{2}{*}{Target Model} & \multirow{2}{*}{Attack Method} & \multicolumn{4}{c}{Metrics}\\ \cmidrule(lr){4-7}
        & & & $FR(\%)$ & $AOA(\%)$ & $AOA^\ast(\%)$ & $AQN$\\
        \midrule
        \multirow{12}{*}{UCF-101} 
        & \multirow{4}{*}{C3D} & PatchAttack~\cite{yang2020patchattack} & 73.3 & 16.9 & 5.7 & 7299\\
        & & Our method (BH) & 65.8 & 8.8 & 2.9 & 10473\\ 
        & & Our method (RL) & \textbf{90.1} & \textbf{7.5} & \textbf{2.5} & \textbf{4273}\\
        & & Our method (Random) & 68.8 & 9.0 & 3.5 & -\\
        \cmidrule(lr){2-7}
        & \multirow{4}{*}{LRCN} & PatchAttack~\cite{yang2020patchattack} & 97.4 & 14.0 & 2.6 & \textbf{1166}\\
        & & Our method (BH) & 97.4 & 8.5 & 2.8 & 1335\\ 
        & & Our method (RL) & \textbf{99.5} & \textbf{5.5} & \textbf{1.0} & 1673\\
        & & Our method (Random) & 97.4 & 8.6 & 2.8 & -\\
        \cmidrule(lr){2-7}
        & \multirow{4}{*}{I3D-Slow} & PatchAttack~\cite{yang2020patchattack} & 92.1 & 14.6 & 4.6 & 2480\\
        & & Our method (BH) & 90.1 & 8.2 & 2.7 & 3468\\ 
        & & Our method (RL) & \textbf{96.5} & \textbf{5.8} & \textbf{1.9} & \textbf{1673}\\
        & & Our method (Random) & 89.6 & 8.2 & 2.8 & -\\
        \cmidrule(lr){1-7}
        \multirow{12}{*}{HMDB-51} 
        & \multirow{4}{*}{C3D} & PatchAttack~\cite{yang2020patchattack} & \textbf{92.2} & 13.5 & 3.5 & \textbf{2500}\\
        & & Our method (BH) & 81.4 & 8.2 & 2.7 & 6358\\ 
        & & Our method (RL) & 91.2 & \textbf{6.4} & \textbf{1.5} & 3122\\
        & & Our method (Random) & 83.3 & 8.8 & 3.2 & -\\
        \cmidrule(lr){2-7}
        & \multirow{4}{*}{LRCN} & PatchAttack~\cite{yang2020patchattack} & 96.9 & 12.1 & 1.6 & 1250\\
        & & Our method (BH) & 94.9 & 8.2 & 2.6 & 1617\\ 
        & & Our method (RL) & \textbf{99.0} & \textbf{4.8} & \textbf{0.7} & \textbf{980}\\
        & & Our method (Random) & 93.9 & 8.0 & 2.5 & -\\
        \cmidrule(lr){2-7}
        & \multirow{4}{*}{I3D-Slow} & PatchAttack~\cite{yang2020patchattack} & \textbf{100.0} & 11.5 & 3.5 & \textbf{760}\\
        & & Our method (BH) & 91.1 & 8.5 & 2.8 & 3453\\ 
        & & Our method (RL) & 99.0 & \textbf{4.8} & \textbf{1.6} & 949\\
        & & Our method (Random) & 98.0 & 7.8 & 2.6 & -\\
        \bottomrule
    \end{tabular}
    \caption{Attack performance on UCF-101/HMDB-51 datasets against C3D/LRCN/I3D-Slow models.}
    \label{result_table}
\end{table*}

Table \ref{result_table} lists the performance comparison against different target models on UCF-101 dataset and HMDB-51 dataset. From the results, we have the following observations. First, compared to PatchAttack, our method that uses BSCs as adversarial patches significantly reduces the occluded area. For all models, the occluded area has been reduced by more than 52\% on both datasets. It is not surprising that BSCs have much smaller occluded areas since compared to a square patch, BSCs are more scattered. Second, compared to BH, RL is more effective in reducing the number of queries. For C3D and LRCN models, the number of queries has been reduced by more than 22\% on both datasets. Besides, RL achieves better performance than random selection under the same query numbers. Third, in most cases, BSCs occlude wider range contents of video than a square patch with the same area and hence increases the fooling rate. Similar results are obtained by conducting experiments on Kinetics-400 \cite{kay2017kinetics} dataset. In summary, using BSCs as adversarial patches decreases the occluded areas and RL helps to achieve a more effective and efficient attack.

Figure \ref{fig-heatmap} shows two examples of adversarial frames generated by our proposed BSC attack method on UCF-101 dataset. In addition, we further visualize the discriminative regions in the video frames for the C3D model with Gradient-weighted Class Activation Mapping (Grad-CAM) \cite{selvaraju2017grad}. From the generated heatmaps, it is clear why the C3D model predicts the input frames as the corresponding correct classes. And embedding the adversarial BSCs into the frame can modify the distribution of the maximum points on the generated heatmap.

To qualitatively evaluate the risks of adversarial patches prone to spot, we use a visual saliency map to show the human-simulated focus area when they take a glance at the image. We compare the BSCs with the square patch, including the original frame as the baseline. Note that both patches occluded the same area of frame for fairness. An example and its saliency map are shown in Figure \ref{fig-saliencymap}. We can see that the square patch can be easily highlighted in the saliency map. This means adversarial patches have a high probability to be spotted at people’s first glance. In contrast, the BSCs are relatively inconspicuous under human observation at first glance. Besides, even if they are detected, BSCs are less likely to arouse people’s suspicion than square patches.

\begin{table}[t]
    \resizebox{.99\columnwidth}{!}{
    \begin{tabular}{l c c c}
        \toprule
        Dataset & Target Model & Type of Patch & $FR(\%)$\\
        \midrule
        \multirow{6}{*}{UCF-101} 
        & \multirow{2}{*}{C3D} & BSC & \textbf{67.9}\\
        & & White Square Patch & 54.2\\
        \cmidrule(lr){2-4}
        & \multirow{2}{*}{LRCN} & BSC & \textbf{81.7}\\
        & & White Square Patch & 75.5\\
        \cmidrule(lr){2-4}
        & \multirow{2}{*}{I3D-Slow} & BSC & \textbf{84.7}\\
        & & White Square Patch & 65.0\\
        \cmidrule(lr){1-4}
        \multirow{6}{*}{HMDB-51} 
        & \multirow{2}{*}{C3D} & BSC & \textbf{70.7}\\
        & & White Square Patch & 59.8\\
        \cmidrule(lr){2-4}
        & \multirow{2}{*}{LRCN} & BSC & \textbf{88.3}\\
        & & White Square Patch & 75.5\\
        \cmidrule(lr){2-4}
        & \multirow{2}{*}{I3D-Slow} & BSC & \textbf{93.9}\\
        & & White Square Patch & 67.3\\
        \bottomrule
    \end{tabular}}
    \caption{Attack performance against the LGS defense.}
    \label{robust_table}
\end{table}

We also evaluate the performance of our attack method against the patch-based defense method - Local Gradient Smoothing (LGS)\cite{naseer2019local}. LGS has shown the best adversarial accuracy on the ImageNet dataset against patch-based attacks among the studied patch defenses to date \cite{chiang2020certified}. In order to evaluate the robustness of adversarial patches with different types, we compare the performance of BSCs and a square patch against LGS defense in terms of the fooling rate. Since the approach is designed for images, we apply the LGS defense operation for each frame in the video. From Table \ref{robust_table}, it is clear that the BSCs are more robust than the square patch against the LGS defense method. Since adversarial training is difficult to apply on videos, an intuitively effective defense method against our BSC attack is to use strong text removal techniques to detect and remove BSCs.

\section{Conclusion}
In this paper, we proposed the BSC attack, a novel black-box adversarial attack against video recognition models. As the meaningful adversarial patch, few BSCs can not only attack the video model easily but also don't arouse people’s suspicion. We formulate the attacking process as an RL problem, where the agent is trained to superimpose BSCs onto the videos in order to induce misclassification. Compared to BH and random selection, RL is much more query efficient and effective. We demonstrated by experiments that compared with the previous PatchAttack, the BSC attack achieves a higher fooling rate while requires fewer queries and occludes smaller areas in the video. Moreover, we also demonstrated that BSCs still have a higher fooling rate than the same area square patch against the LGS defense method.

\section*{Acknowledgement}
This work was supported in part by NSFC project (\#62032006), Science and Technology Commission of Shanghai Municipality Project (20511101000), and in part by Shanghai Pujiang Program (20PJ1401900).

\bibliography{aaai22.bib}

\end{document}